\documentclass[letterpaper]{article}
\usepackage[preprint]{aaai2027}
\usepackage[hyphens]{url}
\usepackage{graphicx}
\usepackage{natbib}
\usepackage{caption}
\usepackage{booktabs}
\usepackage{amsmath,amssymb}

\newcommand{\ours}{RAP-Walk}

\title{When Pairwise Preferences Cycle: A Reliability-Anchored Preference Walk for Video GRPO}

\author{
    Xinrui Lin\textsuperscript{\rm 1},
    Fan Li\textsuperscript{\rm 2,*},
    Shumin Wang\textsuperscript{\rm 1},
    Dehong Kong\textsuperscript{\rm 2},
    Yanyong Zhang\textsuperscript{\rm 1,*}
}
\affiliations{
    \textsuperscript{\rm 1}University of Science and Technology of China, Hefei, China\\
    \textsuperscript{\rm 2}Huawei Noah's Ark Lab
}

\begin{document}

\maketitle

\begingroup
\renewcommand{\thefootnote}{*}
\footnotetext{\raggedright Corresponding authors. Correspondence to: \mbox{Fan Li \textless lifan61@huawei.com\textgreater}, \mbox{Yanyong Zhang \textless yanyongz@ustc.edu.cn\textgreater}.}
\endgroup

\begin{abstract}
Pairwise preference feedback has become a widely adopted supervision signal for reinforcement learning in video generation models, as pairwise comparisons provide a more reliable alternative than pointwise scalar ratings for evaluating the quality of generated videos.
Specifically, we define a preference graph for each GRPO group under a pairwise evaluator, where nodes represent sampled rollouts and edges encode pairwise preferences. Auditing 6,976 on-policy comparison graphs, we find that cyclic preferences are prevalent and persist throughout training. Naive pairwise reward aggregation based on average win rates discards the underlying comparison structures and treats each rollout independently, which may introduce unreliable advantage estimates for certain rollouts and consequently affect reinforcement learning optimization.
Therefore, we propose the \emph{Reliability-Anchored Preference Walk} (\ours{}), a graph-aware advantage aggregation method that strengthens the global preference signal from win rates with reliable relational evidence from cyclic preference graphs through a reliability-guided preference walk. Under matched evaluator, comparison protocol, group size, prompt schedule, and query budget, \ours{} requires no additional evaluator queries beyond the observed comparisons within each rollout group and improves over Pairwise-GRPO across controlled comparisons, transfer settings, and blind human evaluations.
\end{abstract}

\section{Introduction}
Reinforcement learning based methods such as Flow-GRPO and DanceGRPO~\citep{flowgrpo2025,dancegrpo2025} have become important techniques for optimizing video generation models. Early approaches commonly adopt pointwise rewards, which assign an independent scalar score to each generated video. However, video quality is jointly determined by multiple dimensions, including prompt fidelity, visual appearance, motion quality, and temporal consistency, making absolute scores difficult to estimate and calibrate reliably.
Pairwise rewards~\citep{unifiedreward2025,unifiedrewardthink2025,unifiedrewardflex2026} instead evaluate the relative quality of two candidate videos, avoiding the need to define a consistent absolute scoring scale. Consequently, they have recently gained widespread popularity for their improved robustness and interpretability.

Pairwise reward naturally induces a preference graph, where each
rollout is represented as a node and each directed edge encodes both
the preference direction and confidence between two videos. Unlike
pointwise rewards, this graph representation preserves rich relational
information beyond individual scores. However, GRPO ultimately requires
a scalar advantage for each rollout, creating a fundamental
graph-to-advantage mapping challenge:

\textit{How can we reliably reduce a structured preference graph into
the scalar advantages required by GRPO while preserving its relational
information?}

A naive implementation of Pairwise-GRPO~\citep{prefgrpo2025}
addresses this challenge by assigning each rollout a normalized average
win rate against the other rollouts. While this aggregation is simple
and stable, it projects the entire preference graph onto one scalar per
rollout, implicitly treating pairwise outcomes as if they could be
adequately summarized by a globally consistent ordering. In practice,
however, video quality is jointly determined by multiple factors,
including prompt fidelity, visual appearance, motion quality, and
temporal consistency, and pairwise judgments need not be transitive.
For example, an evaluator may prefer video 1 over video 2, video 2 over
video 3, and video 3 over video 1. Such \textbf{preference cycles} should not necessarily
be interpreted as evaluator failures; instead, they may reflect
meaningful trade-offs among different quality dimensions. Naive
win-rate aggregation cannot preserve this structure: comparison graphs
with different opponent relationships and cyclic
patterns may collapse to identical scalar advantages.

This phenomenon is substantial in practice. We audit 6,976 complete
on-policy comparison graphs and find that the proportion of cyclic
groups under the zero-margin definition increases from 79.7\% in the
early stage of training to 89.0\% in the late stage
(Table~\ref{tab:graph-evidence}), while edge clarity simultaneously
decreases. These observations show that preference cycles are not
rare anomalies but a persistent structure in online video
preference optimization.

\begin{figure*}[!t]
\centering
\includegraphics[width=\textwidth]{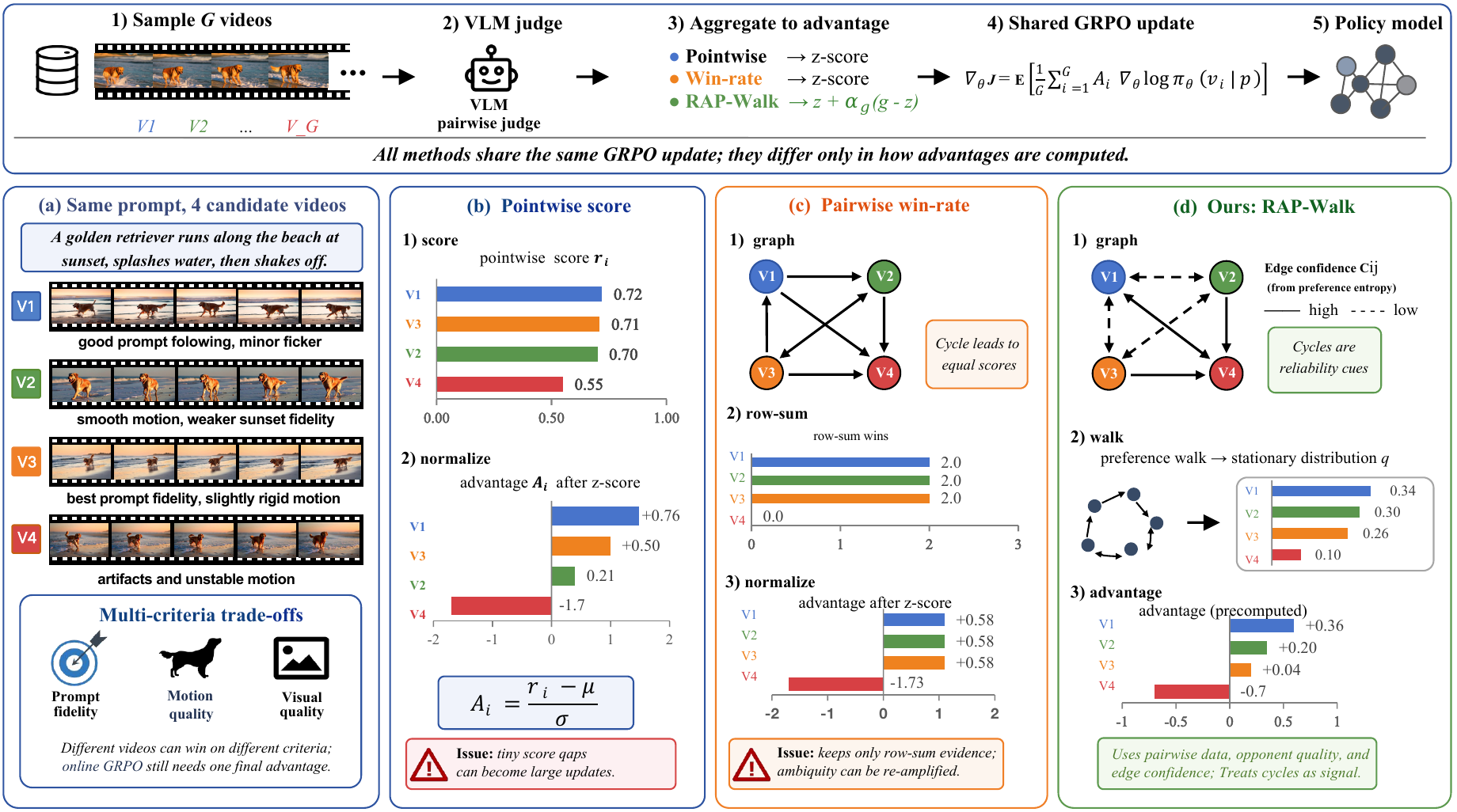}
\caption{Overview of \ours{}. Top: methods share the Pairwise-GRPO pipeline and differ only in advantage construction. Bottom: (a) same-prompt videos exhibit quality trade-offs; (b) pointwise rewards are normalized independently; (c) $v_1$, $v_2$, and $v_3$ form a directed cycle yet have identical row-wise win rates, so win-rate aggregation assigns them identical advantages; (d) \ours{} uses reliable edges and multi-hop paths to correct the win-rate anchor.}
\label{fig:method}
\end{figure*}

To address this graph-to-advantage bottleneck, we propose the
\emph{Reliability-Anchored Preference Walk} (\ours{}), a graph-aware
advantage aggregation method for Pairwise-GRPO.
Rather than replacing average win rate, \ours{} preserves its standardized advantage as a stable optimization anchor and uses the preference graph to construct a controlled structural correction.
Specifically, preference evidence is
propagated from defeated videos to preferred videos, allowing each
video's advantage to depend not only on its direct outcomes but also on
opponent strength and multi-hop comparison paths. To avoid amplifying
ambiguous or strongly non-transitive evidence, the propagation is
modulated by edge reliability and local cyclic inconsistency. The
resulting graph proposal is then combined with the original win-rate
advantage according to the overall reliability of the comparison graph.

In this way, \ours{} treats preference cycles as informative relational
structure rather than errors to be removed, while incorporating graph
information only to the extent supported by edge- and group-level
reliability.
Our contributions are as follows:
\begin{itemize}
\item We reveal that cyclic preference structures naturally emerge in
Pairwise-GRPO for video generation and show that win-rate aggregation
collapses structurally different preference graphs into identical scalar
advantages.
\item We propose \ours{}, a reliability-aware graph aggregation method that preserves the stable win-rate advantage while recovering trustworthy edge and path information from cyclic preference graphs.
\item We validate \ours{} through matched training comparisons, component ablations, transfer across video-generation backbones, and blind human preference evaluations.
\end{itemize}

\section{Related Work}
\paragraph{Reinforcement learning for video generation.}
GRPO was introduced for group-relative policy optimization and later adapted to flow and diffusion generators by Flow-GRPO and DanceGRPO \citep{deepseekmath2024,flowgrpo2025,dancegrpo2025}. Recent video methods study exploration, trust regions, and trajectory-level alignment \citep{sagegrpo2026,tagrpo2026}. These works assume that evaluator feedback has been reduced to scalar rewards or advantages; we study the underexplored mapping from a prompt-local pairwise graph to GRPO advantages.

\paragraph{Preference optimization and aggregation.}
Visual generators now span controllable editing and interleaved synthesis \citep{magiceraser2024,illumex2026}. Preference alignment includes DPO variants for image and video generation \citep{diffusiondpo2023,colorflux2026,videodpo2024,onlinevpo2024,densedpo2025} and scorer-driven GRPO for image editing \citep{hpedit2026}, while UnifiedReward supplies pointwise and pairwise evaluation \citep{unifiedreward2025,unifiedrewardthink2025,unifiedrewardflex2026}. RewardDance selects a reference by pairwise win counts and evaluates new samples against it during ReFL \citep{rewarddance2025}. Closest to our setting, Pref-GRPO uses each sample's within-group win rate as reward \citep{prefgrpo2025}, discarding opponent relations, paths, and cycles. Classical methods use win counts, graph decompositions, or random walks \citep{shahwainwright2015,hodgerank2011,rankcentrality2012}; GPM and NLHF/SPO model cycles with richer representations or game-theoretic objectives \citep{gpm2025,nlhf2024,spo2024}. We instead keep the evaluator and standard GRPO objective fixed, studying only the \emph{preference-graph-to-advantage mapping}, whose magnitudes and reliability directly shape policy updates.

\section{Preliminaries: Pairwise-GRPO}
\label{preliminary}

We first review Pairwise-GRPO, which provides the baseline
formulation and notation used throughout this work. Given a
text prompt $c$, the current policy samples a group of $G$
videos, denoted by $X=\{x_i\}_{i=1}^{G}$. A fixed pairwise
evaluator compares every unordered pair $\{x_i,x_j\}$ in the
group. Throughout this work, we query each pair in both
presentation orders, align the resulting predictions, and
average their soft preference scores. This produces a
reciprocal soft preference matrix
$\mathbf{W}\in[0,1]^{G\times G}$:
\begin{equation}
W_{ij}=P(x_i\succ x_j\mid c),\qquad
W_{ji}=1-W_{ij}\quad (i\ne j),
\label{eq:pairwise-graph}
\end{equation}
with $W_{ii}=0$. Here, $W_{ij}$ quantifies the evaluator's
support for $x_i$ over $x_j$. Accordingly, $\mathbf{W}$ can be viewed
as a weighted directed comparison graph whose nodes are the
sampled videos; under the loser-to-winner convention used
later, $W_{ij}$ represents soft preference evidence directed
from $x_j$ to $x_i$. Bidirectional querying is used to
mitigate presentation-order bias, but it does not impose
transitivity on the resulting comparison graph.

Although the evaluator produces relational feedback, the
GRPO update requires one scalar advantage for each rollout.
Pairwise-GRPO performs this graph-to-scalar reduction by
averaging the off-diagonal entries in each row of $\mathbf{W}$ and
standardizing the resulting win rates within the group:
\begin{equation}
R_i=\frac{1}{G-1}\sum_{j, j\ne i}W_{ij},\qquad
A_i=\frac{R_i-\operatorname{mean}(\mathbf{R})}
{\operatorname{std}(\mathbf{R})+\epsilon},
\label{eq:pairwise-anchor}
\end{equation}
where $\mathbf{R}=\{R_i\}_{i=1}^{G}$ denotes the
win-rate-based rewards, and
$\mathbf{A}=\{A_i\}_{i=1}^{G}$ denotes the corresponding
standardized advantages.
The resulting advantages are used in the standard GRPO objective; the
only modification introduced by RAP-Walk is the construction of the
advantage signal.

This row-mean reduction is simple and incorporates every
direct comparison, but only through the row sums of $\mathbf{W}$.
Consequently, it discards opponent identities and
higher-order relational structure, including comparison
paths and preference cycles.

\section{Reliability-Anchored Preference Walk}
\label{sec:method}

Reliability-Anchored Preference Walk (RAP-Walk) augments the standard Pairwise-GRPO advantage by
constructing a reliability-controlled graph proposal from the preference graph.
The key idea of RAP-Walk is to treat graph propagation as a
controlled correction rather than a replacement of the standard GRPO
signal:

\[
\widetilde{\mathbf A}
=
(1-\alpha_g)\mathbf A+\alpha_g\mathbf g.
\]

Here, $\mathbf A$ preserves the stable win-rate based optimization
signal, while $\mathbf g$ injects relational information recovered from
the preference graph.

\begin{figure}[t]
  \includegraphics[width=\columnwidth]{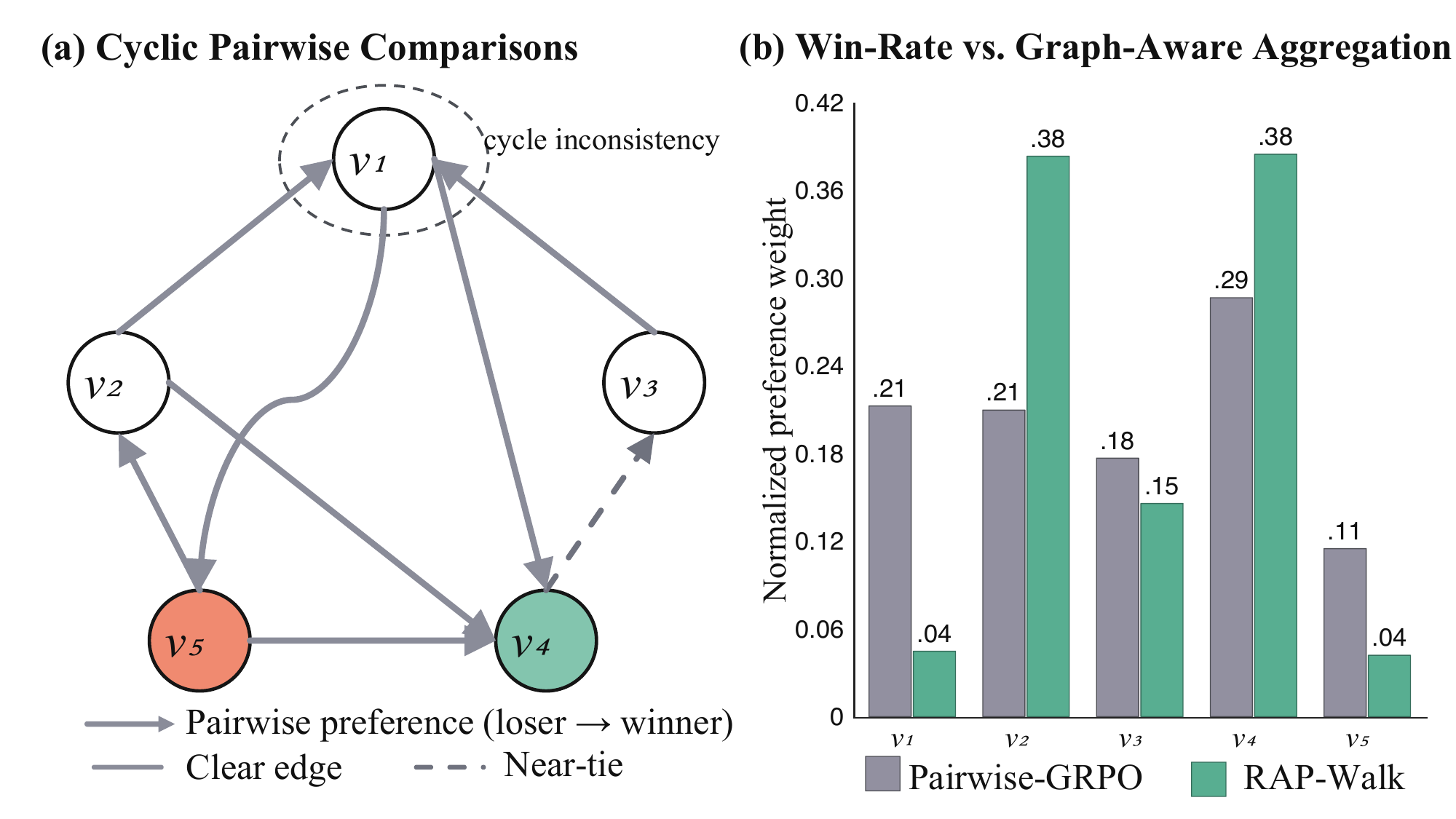}
  \caption{
Five-node illustration of cyclic preference propagation.
(a) A cyclic preference graph with reliability-aware propagation.
(b) Normalized preference credits from row-wise win-rate aggregation
and RAP-Walk. Win-rate aggregation only considers direct comparison
outcomes, while RAP-Walk exploits reliable multi-hop support and
reduces the influence of conflicting local structures.
}
  \label{fig:five-node-cycle}
  \end{figure}

\subsection{Reliability-Aware Graph Proposal}

RAP-Walk views the sampled videos as states of a random walk.
The walk propagates preference mass over the pairwise comparison graph.
The walk moves toward a video $x_j$ when the evaluator prefers $x_j$ over $x_i$.
Through repeated propagation, a video receives more credit when it defeats opponents that themselves receive strong support.
The walk therefore captures opponent strength and multi-step preference paths that are discarded by
row-wise
win-rate aggregation.

This propagation is specified by a row-stochastic transition matrix
$T$. In the simplest form, the transition from video $x_i$ to video $x_j$
is proportional to the probability that $x_j$ is preferred over $x_i$:
\begin{equation}
T^{\mathrm{base}}_{ij}\propto W_{ji},
\qquad i\ne j.
\label{eq:base-transition}
\end{equation}
The use of $W_{ji}$ establishes a loser-to-winner direction: preference
mass leaves the current video $x_i$ and flows toward videos that are more
likely to defeat it.

However, direct preference probabilities may be unreliable:
a near-tie comparison can be ambiguous,
and  clear individual comparisons can form cyclically inconsistent structures.
We therefore introduce a transition-reliability factor $r_{ij}$ that
modulates the base transition using edge clarity and local cyclic
inconsistency. We next define these reliability signals before giving
the complete transition matrix.

\paragraph{Edge clarity.}
We measure the clarity of each pairwise comparison using normalized
Bernoulli entropy:
\begin{equation}
\begin{aligned}
c_{ij}
&=
1-\frac{H_b(W_{ij})}{\log 2},\\
H_b(p)
&=
-p\log p-(1-p)\log(1-p).
\end{aligned}
\label{eq:edge-clarity}
\end{equation}
The clarity $c_{ij}$ approaches zero when $W_{ij}$ is close to $0.5$
and increases as $W_{ij}$ approaches either $0$ or $1$, measuring
the decisiveness of a comparison rather than calibrated correctness.

\paragraph{Cycle inconsistency.}
Edge clarity evaluates individual comparisons but cannot detect
inconsistency that emerges jointly along preference cycles. We first
write the antisymmetric log-odds of each comparison as
$l_{ij}=\log\frac{W_{ij}}{1-W_{ij}}$, where
$l_{ji}=-l_{ij}$. Since the comparison graph is complete, its shortest
nontrivial cycles are triangles. For a triangle $(i,j,k)$, we define its
short-cycle log-odds as
\begin{equation}
L_{ijk}
=
l_{ij}+l_{jk}+l_{ki}.
\label{eq:short-cycle-logodds}
\end{equation}
If comparisons are induced by scalar utilities,
$l_{ij}=u_i-u_j$, then $L_{ijk}=0$;
nonzero value indicates local preference structure
beyond a single scalar ranking.
We summarize the cyclic inconsistency around video $x_i$ by
\begin{equation}
L_i
=
\operatorname*{mean}_{\substack{j<k\\j,k\ne i}}
\left|L_{ijk}\right|.
\label{eq:node-cycle-score}
\end{equation}
We standardize $\{L_i\}_{i=1}^{G}$ across the group in the same manner
as Eq.~\eqref{eq:pairwise-anchor} and, for notational simplicity, reuse
$L_i$ for the standardized score.
A larger $L_i$ therefore indicates
stronger cyclic inconsistency around video $x_i$ relative to the other
videos in the group.

Although cyclic structures contain useful relational information,
high local inconsistency indicates that preference credit should be
propagated more cautiously through that region.

\paragraph{Reliability-weighted transition.}
We combine edge clarity and local cyclic inconsistency through the transition-reliability factor
\begin{equation}
r_{ij}
=
\exp\!\left(-L_j\right)
\left(c_{ij}+\epsilon\right)^{1/4},
\qquad i\ne j.
\label{eq:transition-reliability}
\end{equation}
The factor $\exp(-L_j)$
down-weights nodes in highly non-transitive local structures,
preventing over-amplification of preference credit
supported by conflicting relations while retaining the cyclic evidence.

The fractional power of $c_{ij}$ softly discounts ambiguous comparisons
without suppressing them completely.

To retain the strong first-order signal of Pairwise-GRPO, we use the
standardized win-rate advantage $A_j$ from
Eq.~\eqref{eq:pairwise-anchor} as a direct-support prior for each
destination node. The row-stochastic transition matrix is then
\begin{equation}
T_{ij}
=
\frac{W_{ji}\exp(A_j)r_{ij}}
{\sum_{k\ne i}W_{ki}\exp(A_k)r_{ik}},
\qquad
T_{ii}=0.
\label{eq:transition-matrix}
\end{equation}
The exponential prior provides a first-order preference orientation for
propagation rather than fixing the stationary ranking. The resulting
stationary credit still depends on opponent identities, edge reliability,
and multi-hop paths, which row-wise aggregation does not represent.
Therefore, $\mathbf g$ is not a reparameterization of $\mathbf A$;
it incorporates relational information that is invariant under
row-sum-preserving changes of the preference graph.
Under this construction, preference
mass flows toward videos that are more likely to defeat the current
video, while ambiguous comparisons and locally inconsistent regions
contribute less strongly. Preference cycles are not removed or directly
converted into negative rewards; instead, they reduce the confidence
with which preference credit is propagated through the corresponding
regions.

\subsection{Graph Proposal from Stationary Preference Credit}
We initialize the preference mass uniformly over the sampled videos and
propagate it using the transition matrix:
\begin{equation}
\pi^{(0)}
=
\frac{1}{G}\mathbf{1},
\qquad
\pi^{(t+1)}
=
\pi^{(t)}T.
\label{eq:preference-propagation}
\end{equation}
Repeated propagation approaches a stationary support distribution
satisfying
\begin{equation}
\pi=\pi T.
\label{eq:stationary-distribution}
\end{equation}

The stationary mass $\pi_i$ represents the long-run preference credit
assigned to video $i$. Unlike the row-mean win rate, this credit depends
not only on how often $i$ is preferred, but also on which opponents it
defeats and whether reliable multi-step preference paths lead to it.

A finite comparison graph can
produce an overly
concentrated stationary distribution.
To prevent excessively sharp graph corrections, we define the
normalized entropy
\begin{equation}
\mathcal{H}_{\mathrm{norm}}(p)
=
-\frac{1}{\log G}
\sum_{i=1}^{G}p_i\log p_i.
\label{eq:normalized-entropy}
\end{equation}
We then apply the minimum power-tempering calibration required to
satisfy the entropy floor:
\begin{equation}
q
=
\mathcal{P}_{\xi}(\pi),
\qquad
\mathcal{H}_{\mathrm{norm}}(q)\ge\xi.
\label{eq:entropy-calibration}
\end{equation}
Specifically, $\mathcal{P}_{\xi}$ applies power tempering:
$q_i\propto\pi_i^a$ with $a\in[0,1]$, where the largest exponent
satisfying $\mathcal H_{\mathrm{norm}}(q)\ge\xi$ is selected.
When the stationary distribution already satisfies the entropy floor,
$a=1$ and no calibration is applied.

Finally, we express the calibrated preference credit relative to the
uniform level:
\begin{equation}
g_i
=
Gq_i-1.
\label{eq:graph-proposal}
\end{equation}
Because $\sum_iq_i=1$, the resulting graph proposal satisfies
$\sum_i g_i=0$
, with positive (negative) $g_i$ indicating above (below) uniform graph support.
The proposal $\mathbf{g}=\{g_i\}_{i=1}^G$ is not used directly
as a rollout reward; it provides a graph-aware correction to the
Pairwise-GRPO advantage.

\subsection{Reliability-Anchored Advantage}

The reliability of the graph proposal varies across rollout groups. We
summarize the overall clarity of a comparison graph by averaging its
edge clarity:
\begin{equation}
\bar{c}_g
=
\frac{1}{G(G-1)}
\sum_{i\ne j}c_{ij}.
\label{eq:group-clarity}
\end{equation}
We then map this quantity to a group-level reliability coefficient:
\begin{equation}
\alpha_g
=
\operatorname{proj}_{[\alpha_-,\alpha_+]}
\left(\bar{c}_g\right),
\label{eq:group-reliability}
\end{equation}
where $[\alpha_-,\alpha_+]\subseteq[0,1]$ is a fixed trust interval.
When most comparisons are near ties, $\alpha_g$ is small and limits
 the graph proposal's influence; clearer comparisons increase $\alpha_g$,
 with $\alpha_+$ bounding its maximum contribution.

Using the standardized Pairwise-GRPO advantage $\mathbf{A}$ as the optimization
anchor, we define the corrected advantage as
\begin{equation}
\widetilde{\mathbf{A}}
=
\mathbf{A}+\alpha_g(\mathbf{g}-\mathbf{A})
=
(1-\alpha_g)\mathbf{A}+\alpha_g \mathbf{g}.
\label{eq:anchored-core}
\end{equation}

The corrected advantage is then standardized within the rollout group
(Eq.~\eqref{eq:pairwise-anchor}) before being passed to the standard
GRPO objective. All compared aggregation methods use the same
 standardization; policy-ratio, clipping, and regularization remain unchanged.

\paragraph{Illustrative example.}
Figure~\ref{fig:five-node-cycle} contrasts RAP-Walk with row-wise
  win-rate aggregation on a five-node cyclic graph.

\section{Experiments}
\label{sec:experiments}

\subsection{Experimental Setup}

\paragraph{Datasets and models.}
Following the data protocol of DanceGRPO~\citep{dancegrpo2025}, we curate 10,000 training prompts and an additional disjoint set of 1,000 evaluation prompts from VidProM~\citep{vidprom2024}. The same training and test splits are used for all methods. We employ the Wan family~\citep{wan2025,wan22model2025} as our foundation models, validating our method on the 1.3B and 14B variants of Wan2.1, as well as Wan2.2-TI2V-5B. We instantiate the reward evaluator with frozen Qwen3-VL models (30B and 235B)~\citep{qwen3vl2025}.

\paragraph{Implementation details.}
We train all models for 200 steps using LoRA and AdamW with a learning rate of $3\times10^{-4}$ and group size $G=12$.
We generate 41-frame videos at 480p for Wan2.1 and 720p for Wan2.2, using 30 denoising steps and a guidance scale of 3.
Each evaluator uniformly samples 10 frames from every video. Within each generator, all compared methods share the initialization and training budget.
For Wan2.2-TI2V-5B, we use the TI2V checkpoint in text-to-video mode.

\paragraph{Evaluation metrics.}
For quantitative evaluation, we report six VBench dimensions~\citep{vbench2023}: subject consistency (Subj.), background consistency (Back.), motion smoothness (Smooth.), aesthetic quality (Aesth.), imaging quality (Image), and dynamic degree (Dyn.). We additionally use VideoReward~\citep{videoreward2025}, an evaluation model independent of the Qwen3-VL evaluator used during training, to measure visual quality (VQ), motion quality (MQ), and text alignment (TA); higher scores indicate better performance. Dynamic degree measures motion magnitude rather than perceptual motion quality.

\subsection{Preference Cycles Identification}

Table~\ref{tab:graph-evidence} summarizes 6,976 on-policy preference graphs logged during Pairwise-GRPO training over three stages. Cycle statistics at margin $m$ retain directed edges with $W_{ij}>.5+m$; near ties satisfy $|W_{ij}-.5|\leq .1$.

\begin{table}[t]
\centering
\scriptsize
\setlength{\tabcolsep}{1.35pt}
\renewcommand{\arraystretch}{1.03}
\begin{tabular}{@{}lrrrrrrr@{}}
\toprule
& Node & \multicolumn{2}{c}{Edge} & Triangle & \multicolumn{3}{c}{Groups with a cycle} \\
\cmidrule(lr){2-2}\cmidrule(lr){3-4}\cmidrule(lr){5-5}\cmidrule(l){6-8}
Progress & $L_i$ ratio & Clarity & Tie@$0.10$ & Dens.@$0$ & @$0$ & @$0.025$ & @$0.05$ \\
\midrule
$0$--$33\%$ & $1.34\times$ & 0.261 & 52.6\% & 4.66\% & 79.7\% & 59.5\% & 24.8\% \\
$33$--$66\%$ & $1.31\times$ & 0.162 & 67.6\% & 5.30\% & 87.6\% & 62.1\% & 18.0\% \\
$66$--$100\%$ & $1.32\times$ & 0.132 & 67.3\% & 5.22\% & 89.0\% & 65.0\% & 19.3\% \\
\bottomrule
\end{tabular}
\caption{Preference-graph structure over training. The $L_i$ ratio is the group-averaged maximum-to-mean ratio of unstandardized node inconsistency.}
\label{tab:graph-evidence}
\end{table}

At $m=0$, the any-cycle rate rises from 79.7\% to 89.0\%, while cyclic-triangle density remains near 5\%. At $m=.05$, the rates are 24.8/18.0/19.3\%; meanwhile, clarity falls from 0.261 to 0.132 as the near-tie share rises from 52.6\% to 67.3\%. In fixed-graph replay, holding $W$ fixed and toggling cycle modulation shifts the lowest-$L_i$ third by $+0.616$ and the highest by $-0.544$, isolating its effect from on-policy policy drift.

\subsection{Main Results of RAP-Walk}
\paragraph{Comparison across evaluator capacities.}
\begin{table*}[t]
\centering
\scriptsize
\begin{minipage}[t]{0.492\textwidth}
\centering
\textbf{(a) Qwen3-VL-30B evaluator}\par
\begin{tabular}{@{}l*{9}{@{\hspace{1.50pt}}r}@{}}
\toprule
& \multicolumn{6}{c}{VBench} & \multicolumn{3}{c}{VideoReward} \\
\cmidrule(lr){2-7}\cmidrule(l){8-10}
\textbf{Method}
& Subj. & Back. & Smooth. & Aesth. & Image & Dyn.
& VQ & MQ & TA \\
\midrule
Pretrained (no LoRA)
& 94.60 & 95.50 & 98.46 & 56.73 & 62.37 & \underline{50.00} & 1.988 & 0.552 & 0.530 \\
\addlinespace[1pt]
Pointwise GRPO
& 97.58 & 96.67 & 99.31 & 59.82 & 66.94 & 23.80 & 2.091 & \textbf{0.732} & 0.686 \\
\addlinespace[1pt]
\multicolumn{10}{@{}l}{\textbf{Pairwise aggregation baselines}} \\
\quad Rank Centrality
& 96.94 & 95.40 & 98.71 & 58.98 & 68.79 & \textbf{62.60} & 1.911 & 0.165 & 0.862 \\
\quad Bradley--Terry
& 97.27 & 96.84 & 99.16 & 61.30 & \textbf{69.23} & 37.40 & 2.049 & 0.325 & 1.141 \\
\quad Win rate
& \underline{97.76} & \underline{97.17} & \textbf{99.52} &\underline{62.30}  & 67.78 & 10.00 &  \underline{2.126}  & 0.299 & \underline{1.142} \\
\midrule
\textbf{\ours{}}
& \textbf{98.37} & \textbf{98.13} & \underline{99.42} & \textbf{63.56} & \underline{69.22} & 27.60 & \textbf{2.325} & \underline{0.593} & \textbf{1.289} \\
\bottomrule
\end{tabular}
\end{minipage}
\hfill
\begin{minipage}[t]{0.492\textwidth}
\centering
\textbf{(b) Qwen3-VL-235B evaluator}\par
\begin{tabular}{@{}l*{9}{@{\hspace{1.15pt}}r}@{}}

\toprule
& \multicolumn{6}{c}{VBench} & \multicolumn{3}{c}{VideoReward} \\
\cmidrule(lr){2-7}\cmidrule(l){8-10}
\textbf{Method}
& Subj. & Back. & Smooth. & Aesth. & Image & Dyn.
& VQ & MQ & TA \\
\midrule
Pretrained (no LoRA)
& 94.60 & 95.50 & 98.46 & 56.73 & 62.37 & \underline{50.00} & 1.988 & 0.552 & 0.530 \\
\addlinespace[1pt]
Pointwise GRPO
& \underline{97.00} & \textbf{96.65} & 99.16  & 61.60 & 64.41 & 6.00 & 2.111 & \textbf{0.759} & 0.716 \\
\addlinespace[1pt]
\multicolumn{10}{@{}l}{\textbf{Pairwise aggregation baselines}} \\
\quad Rank Centrality
& 95.12 & 94.40 & 98.74 & 59.16 & 68.81 & \textbf{61.40} & 1.863 & 0.140 & 0.876 \\
\quad Bradley--Terry
& 96.34 & 95.82 & 99.17 & 61.41 & \underline{69.26} & 39.40 &2.059 & 0.358 & 1.162 \\
\quad Win rate
& 96.56 & 95.91 & \textbf{99.53} & \underline{61.94} & 68.24 & 28.60 &  \underline{2.130} & 0.289 & \underline{1.173} \\
\midrule
\textbf{\ours{}}
& \textbf{97.02} & \underline{96.51} & \underline{99.26} & \textbf{63.26} & \textbf{70.09} & 29.80 & \textbf{2.397} & \underline{0.655} & \textbf{1.238} \\
\bottomrule

\end{tabular}
\end{minipage}
\caption{Quantitative comparison on Wan2.1-T2V-1.3B with Qwen3-VL-30B/235B evaluators. Within each evaluator, runs share group size, prompt schedule, and budget; pairwise methods also share the bidirectional comparison protocol and query count. Best/second-best are bolded/underlined.}
\label{tab:main-results}
\end{table*}

Table~\ref{tab:main-results} compares pointwise GRPO~\citep{flowgrpo2025,dancegrpo2025} with four pairwise aggregators: win rate~\citep{prefgrpo2025}, Bradley--Terry~\citep{bradleyterry1952}, Rank Centrality~\citep{rankcentrality2012}, and \ours{}. All trained methods share the evaluator, group size, prompt schedule, and training budget; pairwise methods also share the comparison protocol and query count. \ours{} adds no queries beyond the observed graph.

Relative to win rate, \ours{} improves eight of nine metrics with both evaluators; only smoothness regresses marginally.
Independent VideoReward gains are $+0.199/+0.294/+0.147$ (VQ/MQ/TA, 30B) and $+0.267/+0.366/+0.065$ (235B). Meanwhile, dynamic degree recovers from 10.00/28.60 to 27.60/29.80, with first- or second-place results on every non-dynamic VBench metric. Rank Centrality and Bradley--Terry can retain more motion but lose perceptual quality, suggesting that global graph ranking alone is insufficient without reliability control and anchoring.

\paragraph{Generalization across generator backbones.}
\begin{table}[t]
\centering
\scriptsize
\begin{tabular}{@{}l*{9}{@{\hspace{1.50pt}}r}@{}}
\toprule
& \multicolumn{6}{c}{VBench} & \multicolumn{3}{c}{VideoReward} \\
\cmidrule(lr){2-7}\cmidrule(l){8-10}
\textbf{Method} & Subj. & Back. & Smooth. & Aesth. & Image & Dyn. & VQ & MQ & TA \\
\midrule
\multicolumn{10}{@{}l}{\textbf{Wan2.1-T2V-14B}} \\
Pretrained & 95.43 & 95.92 & 95.92 & 60.68 & 65.47 & \textbf{47.20} & 2.266 & 0.750 & 1.289 \\
Pointwise GRPO & 96.68 & \underline{96.75} & \underline{99.05} & 61.64 & 63.29 & 34.20 & \textbf{2.556} & \textbf{1.051} & 1.211 \\
Win-rate Pairwise & \underline{97.03} & \textbf{96.93} & 99.03 & \underline{62.70} & \underline{66.42} & 36.40 & 2.435 & 0.864 & \underline{1.347} \\
\ours{} & \textbf{97.42} & 96.57 & \textbf{99.11} & \textbf{63.13} & \textbf{67.39} & \underline{46.40} & \underline{2.540} & \underline{0.897} & \textbf{1.585} \\
\midrule
\multicolumn{10}{@{}l}{\textbf{Wan2.2-TI2V-5B}} \\
Pretrained & 96.53 & 96.98 & 98.94 & 57.78 & 65.97 & \underline{40.60} & 2.164 & \underline{0.445} & 1.025 \\
Pointwise GRPO & 96.88 & 97.26 & 99.01 & 58.51 & 66.70 & 38.60 & \underline{2.231} & 0.430 & 1.054 \\
Win-rate Pairwise  & \underline{97.22} & \underline{97.37} & \textbf{99.30} & \underline{59.09} & \underline{67.97} & 38.00 & 2.006 & 0.367 & \underline{1.171} \\
\ours{} & \textbf{97.52} & \textbf{97.63} & \underline{99.27} & \textbf{60.52} & \textbf{68.12} & \textbf{43.00} & \textbf{2.306} & \textbf{0.446} & \textbf{1.306} \\
\bottomrule
\end{tabular}
\caption{Results across generator backbones with a fixed Qwen3-VL-235B evaluator. Best/second-best values are bolded/underlined per generator.}
\label{tab:transfer-matrix}
\end{table}

With matched schedules and budgets, Table~\ref{tab:transfer-matrix} extends the 1.3B setting in Table~\ref{tab:main-results}(b) to two additional generators. Against win rate, \ours{} improves eight of nine metrics on both, with only background consistency (Wan2.1-14B) or smoothness (Wan2.2-TI2V-5B) regressing. The former gains 10.00 dynamic-degree points and 0.238 TA; the latter gains 1.43 aesthetic points and 0.300/0.079/0.135 VQ/MQ/TA. With the evaluator fixed, this repeated pattern supports a backbone-agnostic mapping.

\subsection{Ablation Study}

\begin{table}[t]
\centering
\scriptsize
\begin{tabular}{@{}l*{9}{@{\hspace{1.15pt}}r}@{}}
\toprule
& \multicolumn{6}{c}{VBench} & \multicolumn{3}{c}{VideoReward} \\
\cmidrule(lr){2-7}\cmidrule(l){8-10}
\textbf{Method} & Subj. & Back. & Smooth. & Aesth. & Image & Dyn. & VQ & MQ & TA \\
\midrule
Win-rate Pairwise
& 96.56 & 95.91 & \textbf{99.53} & 61.94 & 68.24 & 28.60 & 2.130 & 0.289 & 1.173 \\
\midrule
\multicolumn{10}{@{}l}{\textbf{Graph proposal construction}} \\
\hspace{0.45em}w/o win-rate prior
& 97.54 & 96.13 & 99.31 & 62.48 & 68.88 & 21.20 & 2.283 & \underline{0.633} & 1.109 \\
\hspace{0.45em}w/o edge clarity
& 96.63 & 96.03 & 99.26 & 61.50 & 66.76 & 30.80 & 2.203 & 0.624 & 1.128 \\
\hspace{0.45em}w/o cycle modulation
& \underline{97.68} & 96.46 & 99.35 & 63.21 & 69.26 & 20.60 & \underline{2.342} & 0.561 & 1.084 \\
\hspace{0.45em}Graph proposal only
& \textbf{97.74}  & \textbf{97.14}& \underline{99.36} &  \textbf{64.08} & \textbf{70.41} & 17.20 & 2.135 & 0.518 & 0.995 \\
\addlinespace[1pt]
\multicolumn{10}{@{}l}{\textbf{Reliability anchoring}} \\
\hspace{0.45em}Fixed group reliability
& 96.47 & 95.69 & 99.11 & 60.24 & 67.61 & \underline{33.20} & 2.045 & 0.350 & 1.054 \\
\hspace{0.45em}Zero reliability floor
& 96.91 & 96.39 & 99.20 & \underline{63.26} & \underline{70.36} & 28.60 & 2.186 & 0.471 & 1.196 \\
\hspace{0.9em}w/o entropy floor
& 96.43 & 95.59 & 99.11 & 61.13 & 69.15 & \textbf{34.40} & 2.144 & 0.325 & \textbf{1.263} \\
\midrule
\textbf{\ours{}}
& 97.02 & \underline{96.51} & 99.26 & \underline{63.26} & 70.09 & 29.80 & \textbf{2.397} & \textbf{0.655} & \underline{1.238} \\
\bottomrule
\end{tabular}
\caption{Ablation study on Wan2.1-T2V-1.3B with Qwen3-VL-235B. Win-rate Pairwise-GRPO is the reference; best/second-best values are bolded/underlined.}
\label{tab:ablation}
\end{table}

Table~\ref{tab:ablation} ablates graph-proposal construction and reliability anchoring. Graph proposal only standardizes $\mathbf g$ directly instead of using Eq.~\eqref{eq:anchored-core}. Fixed group reliability sets $\alpha_g$ to its full-model mean, preserving average correction strength while removing group adaptivity. Zero reliability floor sets $\alpha_-=0$; w/o entropy floor additionally sets $q=\pi$ in this matched zero-floor setting.

\textbf{Graph proposal and anchoring.} Used alone, $\mathbf g$ ranks first or second on all five non-dynamic VBench dimensions, yet trails the full model by 12.60 dynamic-degree points and 0.262/0.137/0.243 VQ/MQ/TA. Thus, graph propagation provides useful relational signal but is better used to correct than replace win rate. Mean-matched fixed reliability gains 3.40 dynamic-degree points but lowers every non-dynamic VBench and VideoReward metric, supporting group-adaptive correction strength.

\textbf{Transition terms and calibration.} Removing the win-rate prior lowers dynamic degree by 8.60 points and all three VideoReward metrics; removing edge clarity lowers aesthetics and imaging by 1.76 and 3.33 points; removing cycle modulation lowers dynamic degree by 9.20 points and all three VideoReward metrics. These changes support the direct-support prior, ambiguity discount, and $L_i$ modulation in Eq.~\eqref{eq:transition-matrix}. With a zero reliability floor, dynamic degree and all three VideoReward metrics decline despite a 0.27 imaging gain. Relative to this matched variant, removing entropy calibration trades $+5.80$ dynamic degree and $+0.067$ TA for losses on all five non-dynamic VBench dimensions plus VQ and MQ. The full configuration therefore gives the strongest VQ/MQ balance, whereas individual ablations can improve dynamic degree or imaging at the expense of broader quality.

\begin{figure*}[t]
\centering
\includegraphics[width=0.94\textwidth]{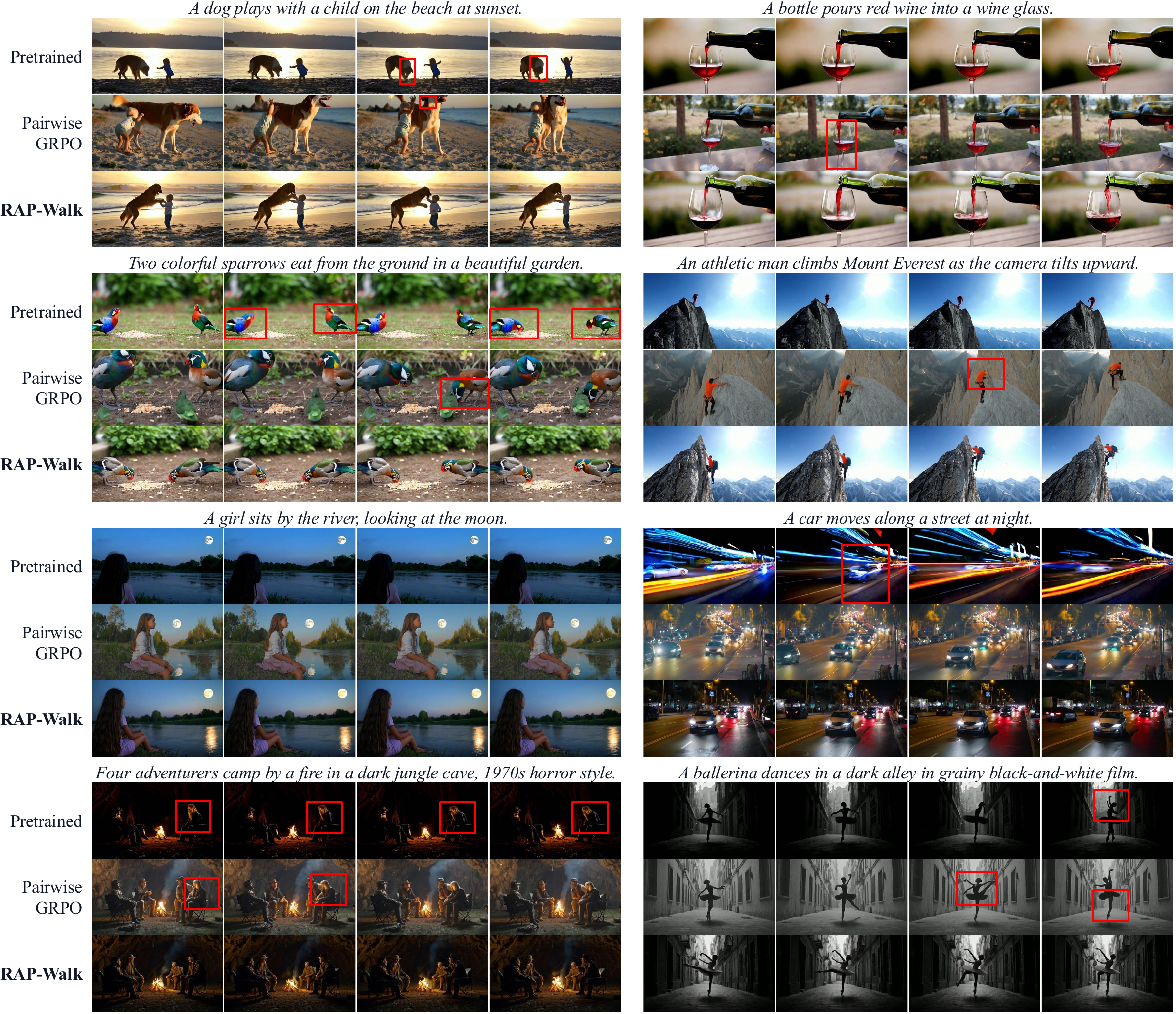}
\caption{Qualitative comparison across eight prompts. Each prompt shows four frames from videos generated by Pretrained, Pairwise GRPO, and \ours{}. Pretrained denotes the original Wan2.1-T2V-1.3B model without LoRA post-training. All methods use matched prompts and initial noise.}
\label{fig:qualitative-results}
\end{figure*}

\subsection{User Study}

\begin{table}[!b]
\centering
\scriptsize
\resizebox{\columnwidth}{!}{\begin{tabular}{@{}l*{10}{@{\hspace{1.3pt}}c}@{}}
\toprule
& \multicolumn{5}{c}{vs. Pairwise} & \multicolumn{5}{c}{vs. Pointwise} \\
\cmidrule(lr){2-6}\cmidrule(l){7-11}
Aspect
& $\gg$ & $>$ & $\approx$ & $<$ & $\ll$
& $\gg$ & $>$ & $\approx$ & $<$ & $\ll$ \\
\midrule
Visual quality
& 27.8 & 36.8 & 2.0 & 20.2 & 13.2
& 46.8 & 31.4 & 1.8 & 7.4 & 12.6 \\
Motion quality
& 17.8 & 31.8 & 14.2 & 22.6 & 13.6
& 32.6 & 26.4 & 14.6 & 9.4 & 17.0 \\
Text alignment
& 29.8 & 23.6 & 18.8 & 13.2 & 14.6
& 44.0 & 20.2 & 13.0 & 6.6 & 16.2 \\
\bottomrule
\end{tabular}
}
\caption{Blind user study (\%). Columns $\gg$, $>$, $\approx$, $<$, and $\ll$ range from strong \ours{} preference to strong baseline preference.}
\label{tab:human-template}
\end{table}

We finally test whether evaluator-independent human judgments corroborate the automatic metrics. In a blind study, 10 raters each evaluated 50 matched-prompt pairs for each baseline comparison on visual quality, motion quality, and text alignment; method identities were hidden and display order was randomized. Against Pairwise/Pointwise GRPO, \ours{} achieves win rates of 64.6\%/78.2\% for visual quality, 49.6\%/59.0\% for motion quality, and 53.4\%/64.2\% for text alignment. Motion is the closest comparison, yet its win--loss margin over win-rate Pairwise-GRPO remains positive at 13.4 points.

These preferences suggest that the gains are not specific to the Qwen evaluator or the reported automatic metrics; the more moderate motion margin also matches the intended conservative correction to the win-rate anchor.

\subsection{Visual Comparison}
Figure~\ref{fig:qualitative-results} compares Pretrained, Pairwise-GRPO, and \ours{} across eight prompts. The baselines show temporal artifacts such as inconsistent dog, child geometry, distorted wine-glass structure, deformed birds, and unstable human or vehicle shapes (red boxes). In contrast, \ours{} preserves object identity and spatial layout across frames while maintaining coherent interactions and motion in the climber, campsite, and ballerina sequences. These results demonstrate improved visual fidelity and temporal consistency without sacrificing scene dynamics or stylistic diversity.

\section{Conclusion}

Pairwise feedback avoids calibrating absolute video-quality scores, yet converting structured preference relations into scalar advantages remains challenging. Our audit of 6,976 on-policy comparison graphs shows that cyclic preference structures emerge during training, indicating that opponent relationships and comparison paths provide information beyond win rates.
We introduced \ours{}, a reliability-anchored graph aggregation method that preserves win-rate advantages as anchors while recovering relational evidence through graph correction. Across matched comparisons, evaluators, backbones, ablations, and blind human evaluations, \ours{} improves over win-rate Pairwise-GRPO without additional evaluator queries.
These results highlight the importance of preserving relational structure when transforming pairwise feedback into optimization signals.

\begingroup
\bibliography{references}
\endgroup

\end{document}